\documentclass[10pt,twocolumn,letterpaper]{article}

\usepackage{cvpr}     

\definecolor{projpink}{RGB}{214,0,143}

\definecolor{cvprblue}{rgb}{0.21,0.49,0.74}

\definecolor{citecolor}{HTML}{2980b9}
\definecolor{linkcolor}{HTML}{c0392b}
\usepackage[pagebackref,breaklinks,colorlinks,citecolor=citecolor,linkcolor=linkcolor]{hyperref}
\usepackage{url}
\usepackage[most]{tcolorbox}
\tcbuselibrary{breakable}
\newtcolorbox{mybreakablepromptbox}[1][]{
  colback=gray!5,
  colframe=black!35,
  boxrule=0.5pt,
  arc=2mm,
  left=6pt, right=6pt, top=5pt, bottom=5pt,
  fontupper=\small\ttfamily,
  title=\textbf{#1},
  breakable,
  enhanced
}
\usepackage{graphicx}
\usepackage{booktabs}
\usepackage[table]{xcolor}   
\usepackage{multirow}

\definecolor{RowOurs}{HTML}{F2F2F2}
\definecolor{RowImg}{HTML}{FEF3EA} 
\definecolor{RowUni}{HTML}{E6F0F9} 
\definecolor{RowVid}{HTML}{EAEAFB} 
\usepackage{tabularx}
\newcolumntype{Y}{>{\centering\arraybackslash}X} 

\usepackage[absolute,overlay]{textpos} 

\def\paperID{*****} 
\def\confName{CVPR}
\def\confYear{2026}

\title{%
\makebox[0pt][l]{%
  \hspace{-1cm}%
  \raisebox{0pt}[0pt][0pt]{%
    \includegraphics[height=2em]{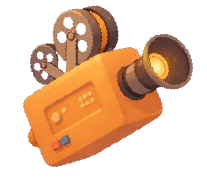}%
  }%
}%
CoF-T2I: Video Models as Pure Visual Reasoners \\ for Text-to-Image Generation}

\author{
Chengzhuo Tong$^{1,2,*}$, Mingkun Chang$^{3,*}$, Shenglong Zhang$^{4}$,
Yuran Wang$^{1,2}$, Cheng Liang$^{2,5}$\\
Zhizheng Zhao$^{1}$, Ruichuan An$^{1}$, Bohan Zeng$^{1,2}$,
Yang Shi$^{1,2}$, Yifan Dai$^{2}$, Ziming Zhao$^{4}$\\
Guanbin Li$^{3}$, Pengfei Wan$^{2}$, Yuanxing Zhang$^{2}$,
Wentao Zhang$^{1,\dagger}$\vspace{0.35cm}\\
Peking University$^{1}$ \quad
Kling Team, Kuaishou Technology$^{2}$ \quad
Sun Yat-sen University$^{3}$\\
\vspace{0.07cm}
Zhejiang University$^{4}$ \quad
Nanjing University$^{5}$\\[0.25em]
Project Page:
\href{https://cof-t2i.github.io}{{\url{https://cof-t2i.github.io}}}
}

\begin{document}
\maketitle
\begin{abstract}
Recent video generation models have revealed the emergence of Chain-of-Frame (CoF) reasoning, enabling frame-by-frame visual inference. With this capability, video models have been successfully applied to various visual tasks (\textit{e.g.}, maze solving, visual puzzles). However, their potential to enhance text-to-image (T2I) generation remains largely unexplored due to the absence of a clearly defined visual reasoning starting point and interpretable intermediate states in T2I generation process.
To bridge this gap, we propose \textbf{CoF-T2I}, a model that integrates CoF reasoning into T2I generation via progressive visual refinement, where intermediate frames act as explicit reasoning steps and the final frame is taken as output.
To establish such explicit generation process, we curate \textbf{CoF-Evol-Instruct}, a dataset of CoF trajectories that model the generation process from semantics to aesthetics. To further improve quality and avoid motion artifacts, we enable independent encoding operation for each frame. Experiments show that CoF-T2I significantly outperforms the base video model and achieves competitive performance on challenging benchmarks, reaching 0.86 on GenEval and 7.468 on Imagine-Bench. These results indicate the substantial promise of video models for advancing high-quality text-to-image generation. 
\end{abstract}

\renewcommand{\thefootnote}{\fnsymbol{footnote}}
\footnotetext{$^*$Equal Contribution\hspace{0.2cm} $^\dagger$Corresponding Author}    
\section{Introduction}

Recent advances in video generation models~\cite{sora,Hunyuan-Video,wan,seedance-1.0} have demonstrated the emergence of zero-shot reasoning behaviors, dubbed \emph{Chain-of-Frame} (CoF) reasoning~\cite{video-zero-shot}. CoF reasoning leverages frame-by-frame generation to iteratively refine scenes, thereby enabling visual inference and unlocking new capabilities in perception, modeling, and manipulation. Benefiting from this, video models such as Veo3~\cite{Veo3} and Sora2~\cite{sora2} have been further extended to a range of downstream visual tasks (\textit{e.g.}, maze solving, visual puzzles)~\cite{video-zero-shot}. 

\begin{figure}
    \centering
    \includegraphics[width=1.0\linewidth]{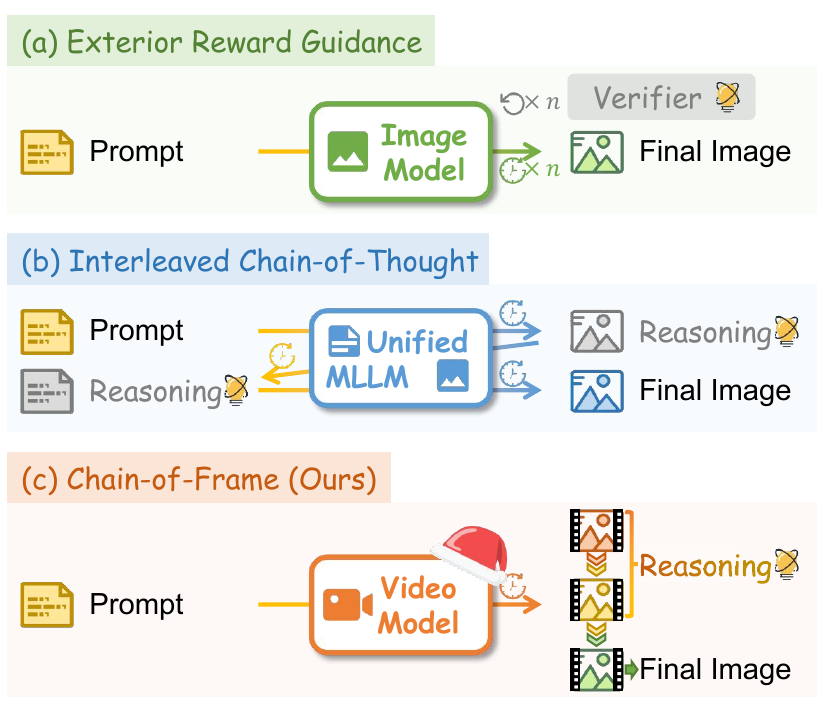}
    \caption{\textbf{Comparison of Inference-time Reasoning Models.} (a) Equipping image models with external verifier. (b) Interleaving textual planning within unified multimodal large language models. (c) \textbf{CoF-T2I}: Our proposed video-based CoF reasoning model.} 
    \label{fig:1}
\end{figure}

\begin{figure*}[!t]
    \setlength{\textfloatsep}{8pt}
    \centering
    \includegraphics[width=1.0\linewidth]{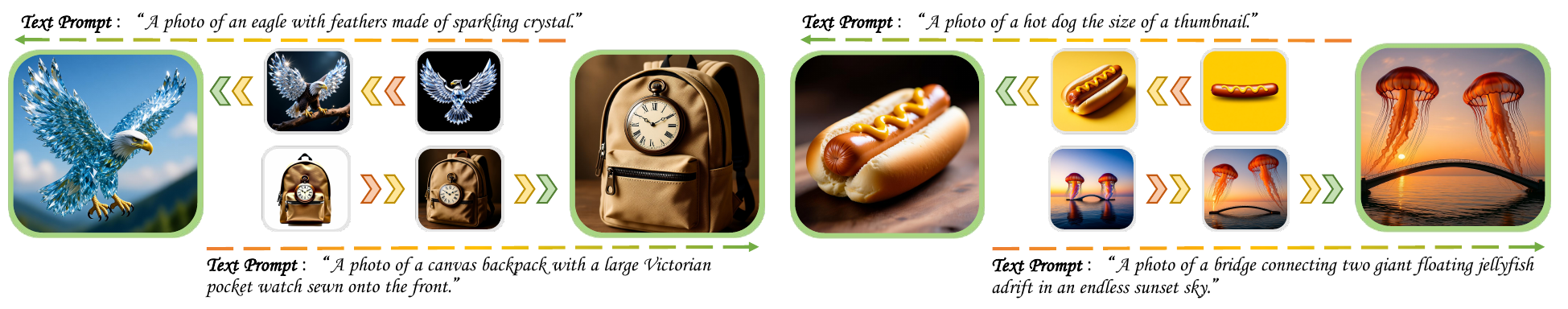}
    \caption{\textbf{Visualizations of CoF-T2I output.} We visualize the reasoning trajectories generated by CoF-T2I. For each example, the final output is shown in large, and the intermediate latent frames are shown in small.}
    \label{fig:visualize_output}
\end{figure*}

Concurrently, the frontier of text-to-image (T2I)~\cite{StableDiffusion-3,gpt4o,show-o,Qwen-Image} generation has shifted toward inference-time reasoning, primarily realized through either employing additional multimodal verifiers to assess image quality~\cite{Image-Generation-CoT,Reflect-DiT} or interleaving textual planning within unified multimodal large language models (MLLMs)~\cite{draco,T2I-R1,TwiG,got,show-o,janus-pro}.
However, these existing reasoning paradigms face two key limitations: first, they rely on frequent modality switching between vision and language (or scalar rewards), making pixel-level correctionsindirect and lossy; second, unified MLLMs lack pretraining on large-scale, purely visual, causally ordered refinement sequences, limiting faithful frame-wise self-correction.
In contrast, video models naturally model the evolution of visual states and refine scenes frame by frame under a strong spatiotemporal prior, making them particularly well-suited as pure visual reasoners to assist high-quality image generation. Still, their potential for enhancing T2I generation remains largely under-explored, primarily due to the lack of a clear visual reasoning starting point and interpretable intermediate states. 
Motivated by this, we raise a fundamental question: \textbf{\emph{Can we use video models as pure visual reasoners to guide high-quality text-to-image generation?}}

To this end, we propose \textbf{\emph{CoF-T2I}}, a model that harnesses the CoF reasoning capability of pretrained video models to enhance T2I generation. Built on a video generation backbone, CoF-T2I reimagines T2I generation as an explicit visual reasoning process. Concretely, given a text prompt, the model generates a compact three-frame sequence, where each frame represents a distinct and progressive reasoning step, from a coarse initial layout to an intermediate refinement, culminating in a high-fidelity final image. To further boost generation quality while suppressing motion artifacts inherent in video backbones, we introduce a frame-wise representation mechanism that allows the model's native video VAE to encode and decode each frame separately, ensuring maximal fidelity. During inference, only the final frame undergoes full decoding and is used as the output image. We present a qualitative comparison between CoF-T2I and other inference-time reasoning models in Figure~\ref{fig:1}.

Internalizing this generation paradigm within video models necessitates vast amounts of structured visual reasoning chains, which are largely absent from existing datasets. To fill this gap, we develop a scalable data curation pipeline and introduce \textbf{\emph{CoF-Evol-Instruct}}, a high-quality dataset comprising 64K CoF sequences that explicitly capture the full T2I generation process from initial semantic composition to final aesthetic refinement.
Our pipeline is carefully designed to produce diverse yet consistently progressive refinement trajectories, providing clear, defect-aware supervision that guides the model from coarse, semantically noisy drafts toward refined, aesthetically coherent final outputs. Together, this dataset and curation pipeline enable effective training of CoF-T2I, allowing the model to internalize strong visual reasoning capabilities.

Experimental results show that CoF-T2I significantly outperforms the base video model and achieves strong performance on challenging benchmarks, attaining a {competitive} score of 0.86 on GenEval and 7.468 on Imagine-Bench. These results highlight the substantial promise of leveraging video foundation models' intrinsic CoF reasoning to advance high-quality T2I generation. The visualization output of CoF-T2I is shown in Figure~\ref{fig:visualize_output}.

In summary, our core contributions are as follows:
\begin{itemize}[topsep=2pt, itemsep=4pt, parsep=1pt, leftmargin=1.5em]
\item \textbf{A novel generation paradigm:} We propose \textbf{\emph{CoF-T2I}}, a text-to-image model that repurposes a video foundation model as pure visual reasoner, generating images via a CoF reasoning process.
\item \textbf{A comprehensive dataset with scalable pipeline: }We introduce \textbf{\emph{CoF-Evol-Instruct}}, a 64K-scale dataset of progressive visual refinement trajectories, built with a scalable quality-aware pipeline.
\item \textbf{Competitive results with extensive validation:} Our extensive experiments show that CoF-T2I substantially outperforms its video backbone and achieves competitive performance on challenging benchmarks, with additional validations confirming its substantial promise.

\end{itemize}

\begin{figure*}[t!]
    \centering
    \includegraphics[width=0.95\linewidth]{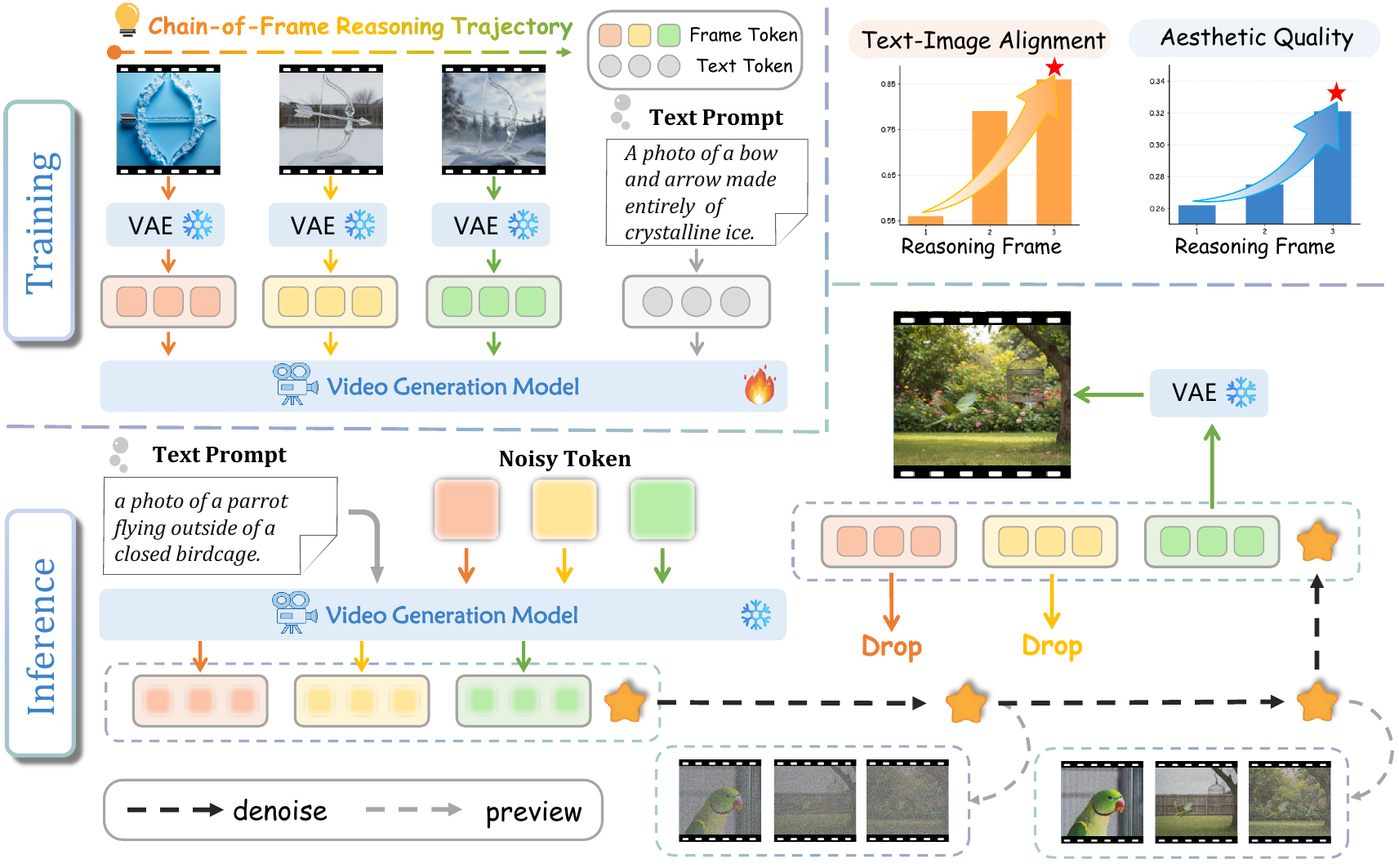}
    \caption{\textbf{Overview of CoF-T2I.} CoF-T2I builds on a video generation backbone, reframing inference-time reasoning for T2I generation as a CoF refinement process. \textbf{Training.} Given a CoF trajectory, we employ a video VAE to encode each frame, and optimize a vanilla flow matching objective. \textbf{Inference.} Starting from noisy initialization, the model denoises to sample a progressively refined reasoning trajectory internalized during training, only the final-frame latent is fully decoded and taken as the output image. \textbf{Quality assessment.} Along the CoF trajectory, text-image alignment and aesthetic quality continue to improve.}
    \label{fig:2}
\end{figure*}
\section{Methodology}
In Section~\ref{sec:preliminary}, we introduce the preliminaries of our work. In Section~\ref{sec:method:generation}, we elaborate on the proposed \emph{CoF-T2I}. In Section~\ref{sec:pipeline}, we present our \emph{CoF-Evol-Instruct} dataset as well as a detailed breakdown of the construction pipeline.

\subsection{Preliminary}
\label{sec:preliminary}

We adopt the Wan2.1 VAE~\cite{wan} to obtain latent representations, which applies causal spatiotemporal compression to raw video frames $ V \in \mathbb{R}^{F \times 3 \times H \times W}$. The first frame is compressed spatially only, while subsequent chunks are conditioned on prior latents for joint spatiotemporal compression. This yields a latent representation with 8 $\times$ spatial downsampling and 4 $\times$ temporal downsampling, resulting in $  F' = \frac{F-1}{4} + 1  $, spatial dimensions $  h = H/8  $, $  w = W/8  $, and channel dimension $C=16$.

Our model adopts Rectified Flow~\cite{flow1-straight-fast-learning,flow2-building-normalizing-flows-stochastic,flow3} to model a straight path from noise ($x_0 \sim\mathcal{N}(0,I)$) to a complex data distribution ($x_1=E(V)$) by learning a velocity field. The interpolated point $x_t=(1-t)x_0+tx_1$ at timestep $t\in [0,1]$ is used to train a scheduler $\mathbf{F_{\theta}}(x_t,t;y,c)$ to predict the direction vector from $x_t$ towards $x_1$, \textit{i.e.}, the velocity field $(x_1-x_0)$. Its training loss minimizes:

\begin{equation}
\begin{aligned}
\mathcal{L}_\theta \,= \,\,&\mathbb{E}_{t \sim p(t),V \sim p_{\text{data}}, x_0 \sim \mathcal{N}(0,I)} \\
&\left[ || \mathbf{F}_{\theta}(x_t, t;y, c) - (x_1 - x_0) ||_2^2 \right] 
\end{aligned}
\end{equation}

where $\boldsymbol{y}$ represents textual conditions and $\boldsymbol{c}$ represents visual conditions. This direct path learning enables high-quality and efficient generation.

\subsection{Image Generation with Video Models}
\label{sec:method:generation}

\noindent\textbf{Overview.}
We propose \textbf{\emph{CoF-T2I}}, a text-to-image foundation model built upon a video generation backbone, as illustrated in Figure~\ref{fig:2}.
We first introduce how video models can be leveraged as \emph{CoF} visual reasoners for T2I generation, then describe the frame-wise latent representation with causal VAE, and finally detail the training and inference procedure of CoF-T2I.

\begin{figure*}[t!]
    \centering
    \includegraphics[width=0.98\linewidth]{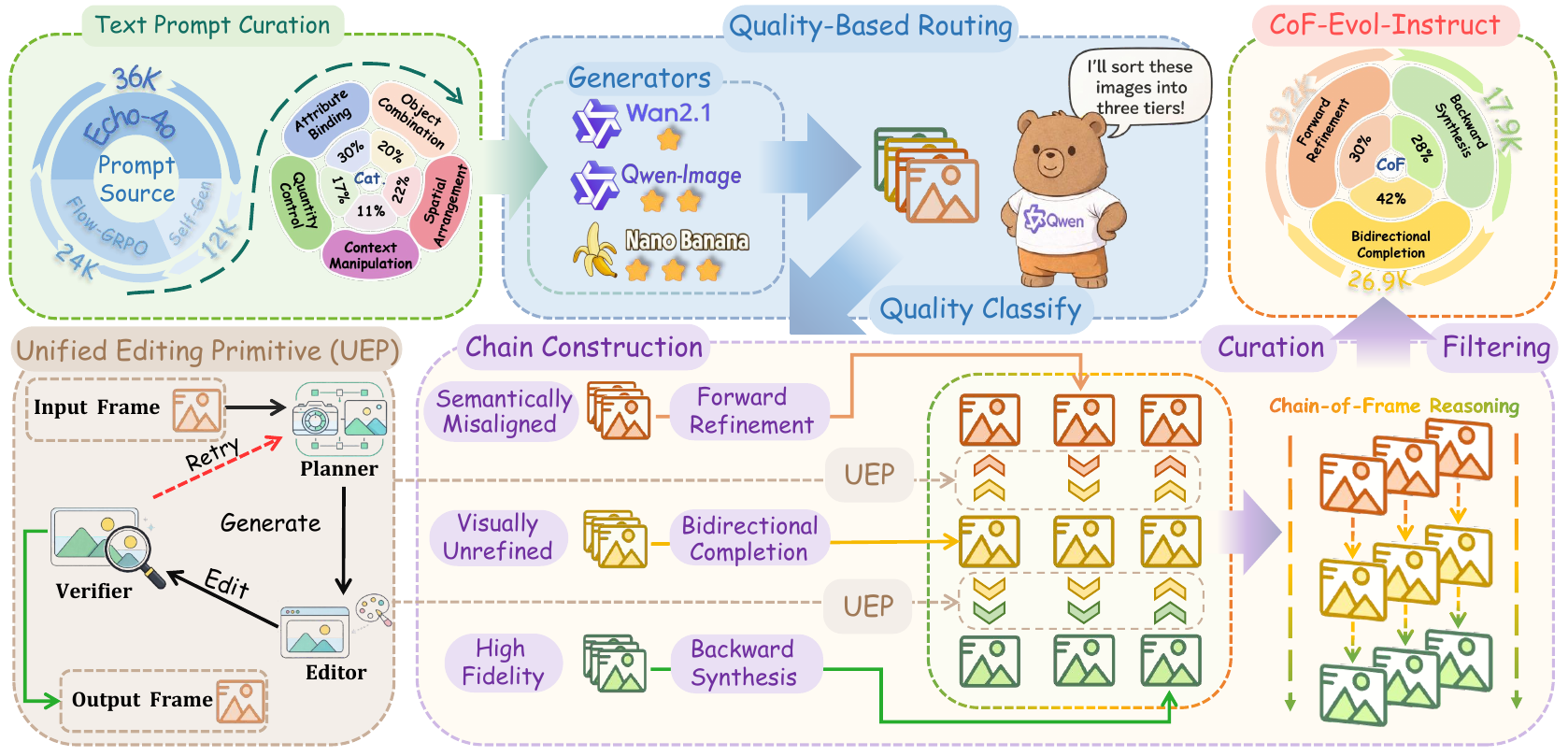}
    \caption{\textbf{Curation Pipeline for CoF-Evol-Instruct.} A \emph{quality-aware construction pipeline} to curate reasoning data. We generate an initial pool of images across diverse distributions and dynamically route valid samples. These images are then expanded into complete CoF sequences through targeted construction strategies. Our pipeline ensures both sample-level diversity and frame-wise consistency.}
    \label{fig:3}
\end{figure*}

\noindent\textbf{Video Models as Visual Reasoners.}
Video foundation models are inherently powerful visual learners and reasoners~\cite{video-zero-shot}, equipped with the natural ability to perform inference-time reasoning through Chain-of-Frame. To harness this  paradigm from a robust video backbone, CoF-T2I redefines the T2I generation process as a structured refinement sequence, enabling frame-wise visual refinement to enhance T2I generation.

Formally, given a text prompt $p$, our method guides the video backbone to produce a sequence of latent representations $z_{1:3}=\{z_1, z_2, z_3\}$. This sequence embodies the CoF logic, evolving from coarse semantics to fine-grained aesthetics. The model achieves this by learning the joint probability distribution of the entire latent trajectory, conditioned by the input prompt:
\begin{equation}
z_{1:3} \sim p_{\theta}(Z_{1:3} | p),
\label{eq:latent_chain_generation}
\end{equation}
Here, $p_{\theta}$ is the probability density over latent sequences learned by the video model. Only the terminal latent state $z_3$, which encapsulates the culmination of the refinement, is projected into the visual space using the decoder $D$ of the native causal VAE, yielding the output image:
\begin{equation}
\hat{I} = F_3 = D(z_3).
\label{eq:latent_to_frame}
\end{equation} 
In this way, the video model iteratively corrects artifacts and enriches details along the logical CoF trajectory, leveraging its sequence-processing architecture without extra textual planning or feedback signals.

\noindent\textbf{Frame-wise Latent Representation.}
\label{cof}
CoF-T2I adopts the video VAE from Wan2.1 to represent frames in a compact latent space. However, the native spatial-temporal compression of VAE may introduce undesired motion artifacts (\emph{e.g.} implicit flow, dynamic inconsistencies). To mitigate this, we employ a frame-wise representation that encodes each frame independently in the latent space: We slide the VAE along the temporal axis of the video and align the input such that the target frame is always compressed within the initial context window, which spans exactly one frame. This design restricts the encoding unit to single-frame granularity, thereby preserving the spatial independence of visual features while enabling high-fidelity compression.

\noindent\textbf{Training and Inference.}
During training, the model is supervised to internalize the frame-wise visual reasoning, leveraging standard flow matching objectives~\cite{flow1-straight-fast-learning}. Concretely, the model predicts the denoising targets for the latent sequence $z_{1:3}=\{z_1, z_2, z_3\}$ corresponding to the CoF sequence $\{F_1, F_2, F_3\}$ and learns to produce later frames that refine earlier ones through end-to-end optimization. At inference,  we generate the full latent sequence starting from Gaussian noise, effectively recovering the internalized generative reasoning via multi-step denoising. Importantly, only the final latent is fully decoded and taken as the final output image $\hat{I}$, while intermediate latents serve solely as internal states for visual reasoning.

\begin{figure*}[t!]
    \centering
    \includegraphics[width=0.98\linewidth]{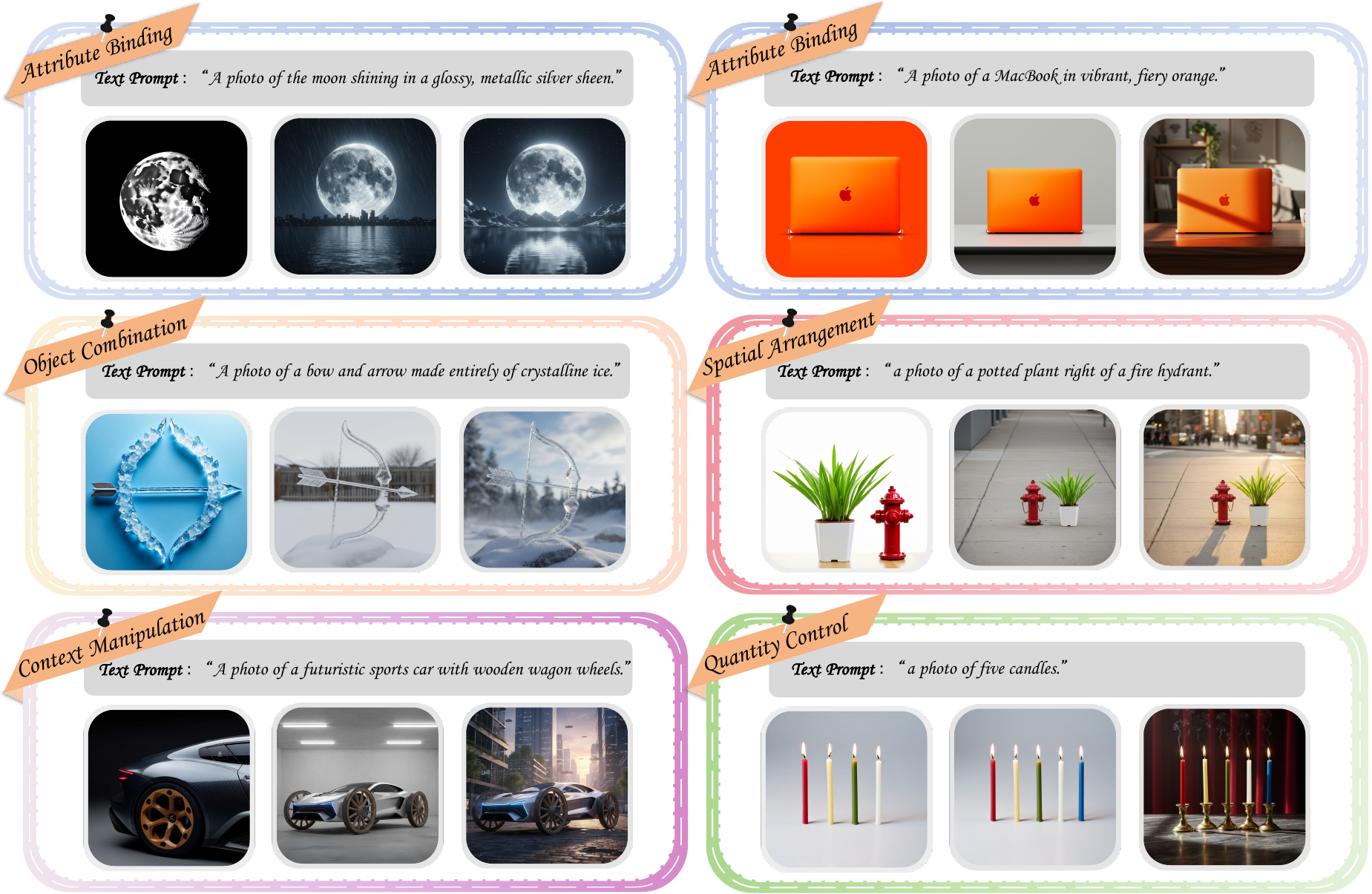}
    \caption{\textbf{Visualization of \emph{CoF-Evol-Instruct} Dataset.} We showcase the prompt and corresponding \emph{CoF} trajectories in our data, including \emph{five} categories: \emph{Attribute Binding}, \emph{Object Combination}, \emph{Spatial Arrangement}, \emph{Context Manipulation}, and \emph{Quantity Control}.}
    \label{fig:4}
\end{figure*}

\subsection{\textbf{\emph{CoF-Evol-Instruct}}}
\label{sec:pipeline}

Constructing diverse CoF trajectories is essential for training video models to perform visual reasoning in T2I tasks, as it allows models to anticipate potential semantic defects and progressively refine aesthetic details on a semantically sound foundation. Below, we first discuss the necessity of constructing this dataset in Section~\ref{no.1}, followed by a detailed description of our quality-aware generation pipeline in Section~\ref{no.2}.

\subsubsection{Necessity of Dataset Construction}
\label{no.1}

Training \emph{CoF-T2I} requires supervision that is both progressive and consistent, reflecting a step-wise refinement process. However, existing T2I datasets~\cite{echo-4o,flux-6m} typically provide single target images, which are insufficient to support the learning of intermediate supervision. Although image editing datasets~\cite{edit1} or multi-step reasoning datasets~\cite{omni-gen2} can be leveraged to construct longer visual generation sequences, they typically suffer from low quality (\textit{e.g.} intermediate degradation, abrupt transitions), thus fail to meet the requirements for strict progressive refinement.

Intuitively, high-quality image generation follows a progressive optimization process: first establishing accurate semantic layout and object placement, followed by the enhancement of aesthetic details and textural fidelity. Motivated by this, we introduce \emph{CoF-Evol-Instruct}, a dataset of three-frame reasoning chains with explicit stage separation and defect-aware construction. In our exploration, we choose a fixed length of three, as it effectively preserves the two key refinement stages (semantic correction and aesthetic improvement) while ensuring consistent causal progression from a defective initial draft to a high-quality final image. This design provides scalable and structured supervision for long-horizon visual refinement in T2I generation.

\subsubsection{Quality-Aware Generation Pipeline}
\label{no.2}

\noindent\textbf{Overview.}
To generate \emph{CoF-Evol-Instruct} at scale while maintaining both sample-level diversity and frame-wise consistency, we propose a quality-aware generation pipeline. Our pipeline begins by sampling a diverse set of images from multiple T2I models spanning different capability tiers. Each image is then assessed for quality and dynamically routed to one of three construction strategies, which subsequently expand it into a complete three-frame CoF sequence: (i) forward refinement, (ii) bidirectional completion, and (iii) backward synthesis. An overview of the proposed pipeline is shown in Figure~\ref{fig:3}.

\noindent\textbf{Multi-model Sampling.}
To ensure a diverse distribution of CoF sequences, we assemble a large collection of text prompts from multiple sources, covering a broad spectrum of scenes, objects, and styles. This includes 24K prompts from~\cite{flow-grpo}, 37K prompts from~\cite{echo-4o}, and 12K self-generated prompts adapted from~\cite{geneval}. After prompt-level deduplication, we retain {68K} unique prompts. Using these prompts, we generate anchor images (\textit{i.e.} the initial images) across several T2I models spanning different capability tiers. In particular, we employ Wan2.1~\cite{wan} as weak tier, Qwen-Image~\cite{Qwen-Image} as medium tier, and Nano-Banana~\cite{nano-banana} as strong tier. To balance quality coverage and avoid being dominated by any single tier, we sample a model tier for each prompt with probabilities $p_\text{weak}=0.25$, $p_\text{medium}=0.5$, $p_\text{strong}=0.25$, yielding {68K} anchor images in total. As a result, we obtain a heterogeneous pool of images that naturally span the coarse-to-fine quality spectrum. Leveraging this diversity in quality levels, we then classify each anchor image to determine the most suitable construction route for expanding it into a full CoF sequence.

\noindent\textbf{Quality-based Routing.}
To exploit this quality diversity, we employ Qwen3-VL-8B~\cite{qwen3-vl} as the quality assessor, which categorizes anchors into three stages based on semantic alignment and aesthetic coherence, producing a three-way classification: \emph{Semantically Misaligned} ($F_1$) for semantically misaligned images regardless of aesthetic performance, \emph{Visually Unrefined} ($F_2$) for semantically accurate but visually unrefined images, and \emph{High Fidelity} ($F_3$) for images achieving both high semantic accuracy and aesthetic coherence. Each classified anchor is then routed to the corresponding construction strategy and expanded into a complete three-frame sequence using the unified editing primitive.
This adaptive routing mechanism ensures optimal starting points for samples across quality spectrum, thereby maximizing dataset coverage and fully utilizing anchors across all quality levels.

\begin{table*}[t]
  \centering
  \caption{\textbf{Performance comparison on GenEval.}
  The best and the second best \emph{Overall} scores are in \textbf{bold} and \underline{underlined}, respectively.}
  \label{tab:geneval}

  \renewcommand{\arraystretch}{1.1}
  \footnotesize

  \begin{tabularx}{\textwidth}{lYYYYYYY}
    \toprule
    \textbf{Model} &
    \textbf{Single Obj.} &
    \textbf{Two Obj.} &
    \textbf{Counting} &
    \textbf{Colors} &
    \textbf{Position} &
    \textbf{Color Attr.} &
    \textbf{Overall}$\uparrow$ \\
    \midrule

    \rowcolor{RowImg}
    \multicolumn{8}{c}{\textbf{\textit{Standard Image Models}}} \\
    \midrule
    SDXL \cite{StableDiffusion-XL}              & 0.98 & 0.74 & 0.39 & 0.85 & 0.15 & 0.23 & 0.55 \\
    SD3-Medium \cite{StableDiffusion-3}   & 0.99 & 0.94 & 0.72 & 0.89 & 0.33 & 0.60 & 0.74 \\
    FLUX.1-dev \cite{flux2024}                   & 0.99 & 0.88 & 0.61 & 0.87 & 0.35 & 0.55 & 0.67 \\
    \midrule

    \rowcolor{RowUni}
    \multicolumn{8}{c}{\textbf{\textit{Unified MLLMs}}} \\
    \midrule
    Janus-Pro-7B \cite{janus-pro} & 0.99 & 0.89 & 0.59 & 0.90 & 0.79 & 0.66 & 0.80 \\
    BLIP3-o 8B \cite{blip3-o}       & --   & --   & --   & --   & --   & --   & \underline{0.84} \\
    OmniGen2 \cite{omni-gen2}                      & 0.99 & 0.92 & 0.77 & 0.90 & 0.82 & 0.70 & 0.80 \\
    BAGEL \cite{bagel}             & 0.99 & 0.94 & 0.81 & 0.88 & 0.64 & 0.63 & 0.78 \\
    BAGEL-Think \cite{bagel}                   & 0.99 & 0.94 & 0.81 & 0.88 & 0.64 & 0.63 & 0.82 \\
    T2I-R1 \cite{T2I-R1}            & 0.99 & 0.91 & 0.53 & 0.91 & 0.76 & 0.65 & 0.79 \\
    \midrule

    \rowcolor{RowVid}
    \multicolumn{8}{c}{\textbf{\textit{Video Models}}} \\
    \midrule
    Wan2.1-T2V-14B \cite{wan}    & 0.92 & 0.63 & 0.57 & 0.69 & 0.18 & 0.31 & 0.55 \\
    \rowcolor{RowOurs}
    \textbf{CoF-T2I (Ours)}        & 0.98 & 0.95 & 0.83 & 0.89 & 0.83 & 0.71 & \textbf{0.86} \\
    \bottomrule
  \end{tabularx}
\end{table*}

\noindent\textbf{Unified Editing Primitive.}
\label{uep}
To construct complete CoF sequences from routed anchors while ensuring cross-frame consistency and stage-specific refinement, we introduce a \emph{unified editing primitive (UEP)} as the shared minimal operation across all strategies. UEP performs controlled, targeted edits for each stage transition (\textit{e.g.}, semantic grounding, aesthetic refinement) while strictly preserving non-target content. We assign each prompt one of five semantic categories: \emph{Attribute Binding}, \emph{Object Combination}, \emph{Quantity Control}, \emph{Spatial Arrangement}, and \emph{Context Manipulation}. Conditioning UEP on the category label narrows editing intent and improves controllability.

UEP is implemented as a closed-loop system with three agents: a planner and verifier powered by Qwen3-VL-32B~\cite{qwen3-vl}, and an editor powered by Qwen-Image-Edit-2509~\cite{Qwen-Image}. Given the current frame, target stage, prompt, transition direction, and category label, the planner generates a minimal editing instruction. The editor applies it while preserving global content. The verifier then evaluates the result against the prompt, target stage, and instruction, outputting a binary success signal $b \in \{0,1\}$. If failed (\ie~$b = 0$), we retry up to $K=3$ times to maintain throughput. This category-conditioned primitive standardizes execution across forward, bidirectional, and backward construction, ensuring prompt-aligned and causally consistent reasoning sequences.

\noindent\textbf{Adaptive Sequence Completion.}
Given anchor images with varying quality, we leverage the proposed UEP and apply three construction strategies to expand each routed anchor into a complete $\{F_1, F_2, F_3\}$ sequence, ensuring progressive refinement regardless of its starting quality stage:

\begin{itemize}[leftmargin=1.2em, labelsep=0.5em, itemsep=3pt, topsep=1pt]
    \item \textbf{Forward Refinement ($F_1 \to F_2 \to F_3$)}. For \emph{semantically misaligned} anchors ($F_1$), which often exhibit semantic defects such as missing objects, incorrect attribute bindings, or implausible spatial layouts. Starting from these coarse generations, we first apply semantic correction via UEP to obtain a prompt-aligned $F_2$. Then, with the semantic layout fixed, we conduct aesthetic enhancement to produce the final $F_3$, improving overall fidelity and lighting coherence.

    \item \textbf{Bidirectional Completion ($F_1 \leftarrow F_2 \to F_3$)}. For \emph{visually unrefined} anchors ($F_2$), which are typically semantically grounded yet lack fine-grained aesthetics. From these intermediates, we expand bidirectionally: Backward via controlled semantic degradation (e.g., weakening attributes, dropping secondary objects), yielding a plausible draft state $F_1$ while preserving global context. Forward via aesthetic refinement to obtain $F_3$, focusing on high-frequency details and visual harmony.

    \item \textbf{Backward Synthesis ($F_1 \leftarrow F_2 \leftarrow F_3$)}. For \emph{high-fidelity} anchors ($F_3$), which represent fully refined and aesthetically superior outputs, we reconstruct backward: first apply aesthetic simplification (reduced sharpness and lighting complexity) to create a semantically grounded $F_2$; then introduce minimal, category-conditioned semantic perturbation (e.g., altering count, degrading attributes) to synthesize a coarse $F_1$. All backward steps are executed and verified with UEP for prompt consistency and inter-frame coherence.
\end{itemize}

\noindent\textbf{Curation and Illustration.}
We apply the proposed pipeline to pre-curated prompts to generate candidate reasoning chains, each paired with its corresponding prompt. After filtering out failed or incomplete samples, the curated collection, named \emph{CoF-Evol-Instruct}, contains 64K high-quality CoF sequences. Representative examples spanning the five categories are presented in Figure~\ref{fig:4}.
The resulting dataset delivers high-quality, diverse, and progressively structured reasoning supervision that is critical for effectively training CoF-T2I. Further details on the dataset construction process are provided in Appendix~\ref{app:data}.


\section{Experiments}
\subsection{Experimental Setting}

\noindent\textbf{Evaluation.}
We evaluate our proposed CoF-T2I on GenEval~\cite{geneval} and Imagine-Bench~\cite{echo-4o}.
GenEval is a widely-used benchmark for object-centric prompt following, focusing on composition, counting, attribute binding, and spatial relations, with automatically verifiable criteria for reliable comparison.
As a complementary setting, Imagine-Bench stresses imaginative prompts that require controlled concept transformations and compositional reasoning (e.g., attribute shifts and hybrid concepts), probing more abstract, concept-level semantics that go beyond literal object-centric instructions.
For both benchmarks, we follow the official evaluation protocols and report the overall and category-wise scores.

\noindent\textbf{Training Details.} CoF-T2I is initialized from the powerful video foundation model Wan2.1-T2V-14B and fine-tuned on our curated CoF-Evol-Instruct 64K dataset for 1,800 steps, utilizing a batch size of 64, a learning rate of $1\text{e-}5$, and weight decay of $1\text{e-}2$. Following the strategy described in Section \ref{cof}, we freeze the VAE and encode each frame of the sequence independently, updating only the unfrozen DiT parameters. Additionally, to mitigate the issue of incomplete subject rendering that is often encountered with default rectangular video aspect ratios (e.g., $480 \times 832$), we standardize our pipeline by resizing all data to $1024 \times 1024$ during training and consistently maintaining this square resolution for inference generation.

\subsection{Main Results}
We report quantitative results in Table~\ref{tab:geneval} and Table~\ref{tab:imaginebench}, respectively.
In our evaluation, we benchmark CoF-T2I against two distinct categories of models: standard image models possessing solely generative capabilities, and unified multimodal models capable of leveraging textual Chain-of-Thought (CoT) for intermediate reasoning.

\begin{table*}[t]
  \centering
  \caption{\textbf{Performance comparison on Imagine-Bench.}
  The best and the second best scores are in \textbf{bold} and \underline{underlined}, respectively.}
  \label{tab:imaginebench}

  \renewcommand{\arraystretch}{1.1} 
  \footnotesize
  \begin{tabularx}{\textwidth}{lYYYYY}
    \toprule
    \textbf{Model} &
    \textbf{Attribute shift} &
    \textbf{Hybridization} &
    \textbf{Multi-Object} &
    \textbf{Spatiotemporal} &
    \textbf{Overall}$\uparrow$ \\
    \midrule
    
    \rowcolor{RowImg}
    \multicolumn{6}{c}{\textbf{\textit{Standard Image Models}}} \\
    \midrule
    SDXL \cite{StableDiffusion-XL}                & 4.420 & 4.930 & 4.500 & 6.320 & 4.970 \\
    SD3-Medium \cite{StableDiffusion-3}         & 5.140 & 6.300 & 6.070 & 5.910 & 5.780 \\
    FLUX.1-dev \cite{flux2024}         & 5.680 & 6.380 & 5.240 & 7.130 & 6.060 \\
    \midrule

    \rowcolor{RowUni}
    \multicolumn{6}{c}{\textbf{\textit{Unified MLLMs}}} \\
    \midrule
    Janus-Pro-7B \cite{janus-pro}        & 5.300 & 6.730 & 6.040 & 7.280 & 6.220 \\
    BLIP3-o 8B \cite{blip3-o}          & 5.800 & 7.060 & 6.440 & 7.080 & 6.510 \\
    OmniGen2 \cite{omni-gen2}            & 5.280 & 6.290 & 6.310 & 7.450 & 6.220 \\
    BAGEL \cite{bagel}               & 5.370 & 6.500 & 6.410 & 6.930 & 6.200 \\
    BAGEL-Think \cite{bagel}         & 6.260 & 7.740 & 6.960 & 7.130 & \underline{6.930} \\
    T2I-R1 \cite{T2I-R1}              & 5.850 & 7.360 & 6.680 & 7.700 & 6.780 \\
    \midrule

    \rowcolor{RowVid}
    \multicolumn{6}{c}{\textbf{\textit{Video Models}}} \\
    \midrule
    Wan2.1-T2V-14B \cite{wan}       & 5.436 & 6.950 & 5.383 & 6.237 & 5.939 \\
    \rowcolor{RowOurs}
    \textbf{CoF-T2I (Ours)} & 6.969 & 8.070 & 7.797 & 7.287 & \textbf{7.468} \\
    \bottomrule
  \end{tabularx}
\end{table*}

On GenEval, CoF-T2I achieves a leading overall score of 0.86, establishing a competitive performance among the compared methods. Notably, our pure visual reasoning approach outperforms strong unified baselines that rely on textual planning, surpassing BAGEL-Think by 0.04 and T2I-R1 by 0.07. This superior performance demonstrates the substantial advantage of our proposed CoF-T2I in generating precise and semantically correct images.
Performance on Imagine-Bench further corroborates the robustness of our model. CoF-T2I delivers a remarkable improvement over the pre-trained Wan2.1 baseline, boosting the overall score from 5.939 to 7.468. Notably, the gain is especially evident under complex composition, where the Multi-Object category reaches 7.797 compared with 5.383 for Wan2.1.
This underscores the strength of CoF-T2I in handling imaginative instructions that require controlled concept transformations and complex compositional reasoning.
Overall, by repurposing video foundation models as intrinsic visual reasoners, we demonstrate that this paradigm is not only viable but possesses immense potential. It offers a promising direction for future text-to-image generation, where the video model’s intrinsic CoF reasoning is leveraged at inference-time to iteratively correct semantics and refine visual details, yielding higher-quality generations.

\begin{table}[t]
\centering
\caption{\textbf{Ablation study on core mechanisms.} \textit{Wan2.1 Base} refers to the Wan2.1-T2V-14B base model used for these comparative experiments. \textit{Target-only SFT} fine-tunes on only the final frame of CoF-Evol-Instruct. \textit{w/o Independent VAE} uses the default causal video-VAE encoding to encode the frame chain continuously.}
\label{tab:ablation}
\setlength{\tabcolsep}{3.5pt}
\resizebox{\columnwidth}{!}{%
\begin{tabular}{lcccccc|c}
\toprule
\textbf{Method} & \textbf{Single} & \textbf{Two Obj.} & \textbf{Counting} & \textbf{Colors} & \textbf{Position} & \textbf{Color Attr.} & \textbf{Overall} \\
\midrule
Wan2.1 Base & 0.92 & 0.63 & 0.57 & 0.69 & 0.18 & 0.31 & 0.55 \\
Target-only SFT & \textbf{0.99} & 0.92 & 0.75 & 0.86 & 0.73 & 0.59 & 0.81 \\
CoF w/o Independent VAE & 0.98 & 0.94 & 0.82 & 0.84 & 0.78 & 0.63 & 0.83 \\
\textbf{CoF-T2I (Ours)} & 0.98 & \textbf{0.95} & \textbf{0.83} & \textbf{0.89} & \textbf{0.83} & \textbf{0.71} & \textbf{0.86} \\
\bottomrule
\end{tabular}}
\end{table}

\begin{table}[t]
\centering
\caption{\textbf{Analysis of the Reasoning Trajectory.} We evaluate GenEval performance across each frame of the reasoning chain.}
\label{tab:steps}
\setlength{\tabcolsep}{4pt}
\resizebox{\columnwidth}{!}{%
\begin{tabular}{lcccccc|c}
\toprule
\textbf{Step} & \textbf{Single} & \textbf{Two Obj.} & \textbf{Counting} & \textbf{Colors} & \textbf{Position} & \textbf{Color Attr.} & \textbf{Overall} \\
\midrule
Frame 1 (Draft)  & 0.95 & 0.58 & 0.62 & 0.56 & 0.49 & 0.16 & 0.56 \\
Frame 2 (Refine) & 0.97 & 0.91 & 0.80 & 0.81 & 0.75 & 0.51 & 0.79 \\
\textbf{Frame 3 (Final)} & \textbf{0.98} & \textbf{0.95} & \textbf{0.83} & \textbf{0.89} & \textbf{0.83} & \textbf{0.71} & \textbf{0.86} \\
\bottomrule
\end{tabular}}
\end{table}

\subsection{Ablation Study}
\label{sec:ablation}

\noindent\textbf{Does intermediate supervision yield additional benefits?}
To answer this question, we examine the contribution of intermediate supervision in our Chain-of-Frame reasoning training data by comparing the full CoF-T2I model with a \textit{Target-Only SFT} variant.
For a fair comparison, CoF-T2I and \textit{Target-Only SFT} are trained under identical settings, including the same training hyperparameters and number of optimization steps.
This variant is fine-tuned using only the final frames $F_3$ from the CoF-Evol-Instruct dataset, with all intermediate reasoning frames removed during training.
As illustrated in Table~\ref{tab:ablation}, the \textit{Target-Only SFT} variant achieves a notable improvement over the Wan2.1 base model, with the overall score rising from 0.55 to 0.81.
However, it still falls short of CoF-T2I at 0.86.
This disparity indicates CoF-T2I benefits not merely from stronger target supervision, but also from explicitly learning the generative trajectory, which leads to superior overall performance.
We observe a consistent trend on Imagine-Bench as well, and report the corresponding ablation results in the Appendix~\ref{sec:more_experiment_details}.

\begin{figure*}
    \centering
    \includegraphics[width=0.96\linewidth]{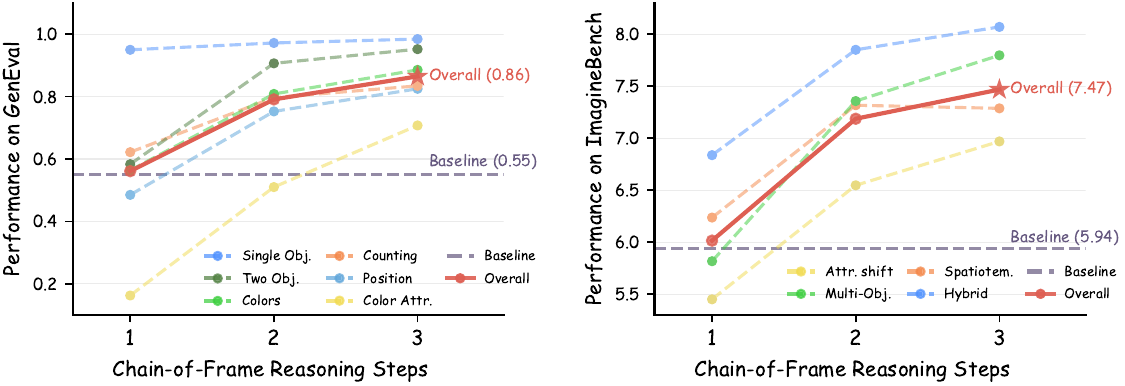}
    \caption{\textbf{Evolution of generation quality across reasoning steps.} We visualize the progressive improvement on both GenEval (left) and Imagine-Bench (right). The results exhibit a general ascending trend in performance scores across the inference steps.}

    \label{fig:reasoning_steps}
\end{figure*}

\noindent\textbf{Analysis of the Reasoning Trajectory.}
To quantify how CoF inference improves generation, we evaluate each intermediate frame in the three-step reasoning chain.
As shown in Table~\ref{tab:steps} and Fig.~\ref{fig:reasoning_steps}, GenEval performance increases monotonically from the draft $F_1$ with a score of 0.56 to the refined $F_2$ at 0.79 and further to the final $F_3$ at 0.86, with consistent gains observed across all sub-tasks.
This steady progression indicates that CoF-T2I enables iterative visual self-correction, in which semantic alignment, perceptual fidelity, and visual coherence are jointly refined at each step, leading to consistent improvements across successive frames.
A similar progression is also observed on Imagine-Bench, suggesting that this refinement behavior generalizes beyond object-centric evaluation settings. Detailed category-wise analyses for Imagine-Bench are provided in the Appendix~\ref{sec:more_experiment_details}.

\noindent\textbf{Robustness Across Model Scales.}
We further examine whether CoF-T2I yields consistent benefits across varying model capacities. 
As shown in Table~\ref{tab:scaling}, applying the same CoF-T2I training recipe to Wan2.1-T2V backbones at 1.3B and 14B parameters stably improves GenEval, indicating the learned frame-evolution trajectory is effective regardless of scale.
Notably, the relative gain is more pronounced on the 1.3B model, while the 14B variant still benefits substantially.
These results validate the robustness of the CoF reasoning paradigm underlying CoF-T2I, demonstrating its effectiveness across diverse model configurations.

\begin{table}[t]
\centering
\caption{\textbf{Robustness of CoF-T2I across model scales.} 
We report GenEval overall scores and absolute gains across 1.3B and 14B scales. \textit{Wan2.1 Base} refers to the Wan2.1-T2V backbone.}

\label{tab:scaling}
\resizebox{\columnwidth}{!}{%
\begin{tabular}{llcc}
\toprule
\textbf{Model Size} & \textbf{Method} & \textbf{Overall Score} & \textbf{Improvement} \\
\midrule
\multirow{2}{*}{1.3B} & Wan2.1 Base & 0.22 & - \\
 & \textbf{CoF-T2I (Ours)} & \textbf{0.79} & \textbf{+0.57} \\
\midrule
\multirow{2}{*}{14B} & Wan2.1 Base & 0.55 & - \\
 & \textbf{CoF-T2I (Ours)} & \textbf{0.86} & \textbf{+0.31} \\
\bottomrule
\end{tabular}}
\end{table}

\noindent\textbf{Necessity of Independent Frame Encoding.}
We further analyze the impact of the encoding strategy in Table~\ref{tab:ablation}.
Specifically, we compare our independent frame encoding against a \textit{w/o Independent VAE} variant that utilizes the default causal video VAE encoding under the same training configuration.
The inferior performance of this variant with a score of 0.83 suggests that treating reasoning steps as distinct and independent visual states is essential for decoupling the refined output from the defective draft.
In contrast, continuous encoding might introduce unnecessary temporal dependencies that entangle reasoning steps, potentially limiting the effectiveness of visual correction.
The implementation details of the continuous video VAE encoding are provided in Appendix~\ref{sec:more_experiment_details}.

\section{Conclusion}
In this work, we introduce \textit{\textbf{CoF-T2I}}, a text-to-image foundation model that repurposes pretrained video generation backbones as \emph{pure visual reasoners}. CoF-T2I performs multi-step frame evolution at inference time, progressively correcting semantic errors and refining perceptual quality to produce higher-fidelity generations without relying on textual planning.
We also curate CoF-Evol-Instruct with specialized pipelines, providing step-wise supervision signals for progressive visual refinement.
Extensive experiments on GenEval and Imagine-Bench demonstrate strong improvements over the base model and competitive performance on challenging prompt-following and compositional reasoning settings. We hope this work motivates future exploration of video-derived temporal reasoning mechanisms for more capable and controllable text-to-image generation.

{
    \small
    \bibliographystyle{ieeenat_fullname}
    \bibliography{main}
}

\clearpage
\setcounter{page}{1}

\appendix

\section*{Appendix Overview}

\begin{itemize}
    \item Section~\ref{app:related}: Related Work.
    \item Section~\ref{app:data}: More Dataset Details.
    \item Section~\ref{sec:more_experiment_details}: More Experiment Details.
    \item Section~\ref{app:limit}: Limitations and Future Work.
    \item Section~\ref{app:qual}: Qualitative Examples.
\end{itemize}

\section{Related Works}
\label{app:related}

\textbf{Reasoning in Visual Generation.}\hspace{0.4em} While models like Stable Diffusion 3~\cite{StableDiffusion-3} and FLUX.1~\cite{flux.1} show strong generation, their single-step text-to-image (T2I) mapping struggles with complex logic. Recent studies~\cite{scaling-llm-test-time,inference-time-scaling} indicate that scaling test-time computation effectively improves performance. Building on this, numerous works propose employing additional models to guide image generation~\cite{Image-Generation-CoT,Reflect-DiT,reflect-perfect}. For instance, Image-Gen-CoT~\cite{Image-Generation-CoT} introduces PARM as a reward model for test-time verification, while ReflectionFlow~\cite{reflect-perfect} provides natural language feedback via external vision language models. Unlike external-signal-reliant approaches, another paradigm~\cite{ulm-trans-fusion,janus,show-o,janus-pro,bagel} uses a unified understanding-generation language model to decompose single-step mapping into interleaved textual reasoning and visual synthesis~\cite{T2I-R1,interleaving-reasoning-better-text-to-image,think-with-generated-images,TwiG,draco}. Specifically, TwiG~\cite{TwiG} interleaves textual reasoning for on-the-fly guidance and Draco~\cite{draco} explores a path of pure visual reasoning, utilizing pre-planned visual drafts to guide the final generation.

\textbf{Zero-Shot Reasoning in Video Models.}\hspace{0.4em} With the rapid development of video generation models~\cite{sora,Veo3,Hunyuan-Video,wan,seedance-1.0}, new avenues for pure visual reasoning are emerging. The quest for step-by-step reasoning, pioneered by the CoT~\cite{CoT1,llm-zero-shot,CoT2}  paradigm in language models, has begun exploring a visual counterpart~\cite{COF1,Image-Generation-CoT}. Chain-of-Frame (CoF) reasoning, introduced by Wiedemer et al.~\cite{video-zero-shot}, first showed video models' ability to tackle complex visual tasks. Like CoT, CoF recasts visual challenges into spatiotemporal progressions, where each frame acts as an intermediate step of visual thought. Based on this, a surge of research~\cite{mme-cof,reason-via-video,V-reason-benchmark,image-to-video-models,veo-benchmark,video-model-benchmark} has focused on systematically evaluating CoF's emergent capability to quantify video models' strengths and limitations across spatial, physical, and logical tasks. These efforts indicate CoF's potential for complex visual reasoning beyond language-mediated approaches. But applying its pure visual reasoning to guide and optimize high-quality image generation remains largely under-explored.

\section{More Dataset Details}
\label{app:data}

\subsection{Details on prompt categorization.}
\label{app:categorization}

To make semantic-stage edits in UEP targeted and controllable, we categorize each prompt by its \emph{primary constraint} (\textit{i.e.}, the dominant factor that must be changed to satisfy the prompt while keeping other content invariant.) Here, we define five categories in detail:

\begin{itemize}[leftmargin=1.2em, labelsep=0.5em, itemsep=3pt, topsep=1pt]
    \item \textbf{Attribute Binding.} Constraints that bind {intrinsic, attributes} to an object, (\textit{e.g.} color, shape/size, material). The object identity and overall scene remain the same, only the specified property changes.
    \item \textbf{Object Combination.} Constraints that concerns {multi-object composition}, including the co-existence of multiple objects or creative combinations (\textit{e.g.}, unusual juxtaposition or hybrid concepts). The object identity is object-level composition.
    \item \textbf{Quantity Control.} Constraints that require a specific number of instances (counts or count-sensitive descriptions). The primary variable is the number of objects.
    \item \textbf{Spatial Arrangement.} Constraints that specify {relative spatial relations} among objects (\textit{e.g.}, left/right, above/below, in/on, front/behind). Object identities are preserved, but the prompt is satisfied only when their spatial configuration matches the relation.
    \item \textbf{Context Manipulation.} Constraints dominated by {global context} that cannot be reduced to one object, such as temporal cues (era/time mismatch) or environmental settings (background/location/scene condition). Satisfying this requires modifying scene-level context.
\end{itemize}

\subsection{Implementation Details.}

\paragraph{Quality-based Routing.}
We leverage {Qwen3-VL-7B}~\cite{qwen3-vl} to perform \emph{Quality-based Routing} for anchor images. Given a prompt and its generated image, we ask the model to judge whether the image is (i) semantically incorrect, (ii) semantically correct but aesthetically under-refined, or (iii) both semantically correct and aesthetically refined. 
Concretely, we perform a three-way classification into: \emph{Semantically Misaligned ($F_1$)}, \emph{Visually Unrefined ($F_2$)}, and \emph{High Fidelity ($F_3$)}. 
$F_1$ correspond to images with noticeable semantic violations (e.g., missing objects, wrong attribute bindings, incorrect relations) regardless of visual fidelity; $F_2$ are semantically aligned but still visually unrefined (e.g., lacking texture, lighting, or realism); and $F_3$ satisfy both semantic alignment and aesthetic coherence, serving as high-quality terminal states for chain construction, the prompt we use for classification is:

\begin{mybreakablepromptbox}[Quality Assessing Prompt Template]

You are a strict image quality assessor. \\
\\
Input:\\
- PROMPT: \{prompt\} \\ 
- IMAGE: [Image Input] \\
\\
Evaluate based on Semantic Alignment and Aesthetic Quality: \\
1. Semantically Misaligned (F1): Major semantic identity errors (e.g., wrong objects), regardless of aesthetics. \\
2. Visually Unrefined (F2): Semantically correct, but low aesthetic quality (blur, distortion, bad lighting). \\
3. High Fidelity (F3): Semantically correct AND high aesthetic quality. \\
\\
Output Format (JSON-compatible):
\begin{verbatim}
{
  "label": "F1" | "F2" | "F3",
  "analysis": "strict reasoning based on 
  the definitions above",
}
\end{verbatim}
\end{mybreakablepromptbox}

\subsection{Unified Editing Primitive} 

The Unified Editing Primitive (UEP) serves as the core modular operation in our quality-aware pipeline, enabling consistent, category-conditioned stage transitions across all three construction routes. UEP is implemented as a closed-loop agent system comprising three components powered by Qwen-family models: a planner (Qwen3-VL-32B~\cite{qwen3-vl}), an editor (Qwen-Image-Edit-2509~\cite{Qwen-Image}), and a verifier (Qwen3-VL-32B~\cite{qwen3-vl}).

\noindent{\textbf{Prompt Designs.}}
To achieve precise control, we design role-specific system prompts for every component of UEP for planner and verifier. We offer simplified templates:
\begin{mybreakablepromptbox}[Planner Prompt Template]

You are a visual reasoning planner for image editing.\\

Input Context:\\
- CURRENT IMAGE: [current frame]\\
- PROMPT: \{prompt\}\\
- STAGE: \{stage\} (semantic correction OR aesthetic refinement)\\
- DIRECTION: \{direction\} (forward OR backward)\\
- CATEGORY: \{category\} (one of [Attribute Binding, etc.])\\
- Edit Hint: \{hint\} (based on CATEGORY) \\
- PREVIOUS FRAME: [prior frame, optional]\\
Editing Logic:\\
Generate a targeted editing instruction. If DIRECTION is forward, fix semantic discrepancies in the CATEGORY or enhance aesthetic realism. If DIRECTION is backward, explicitly introduce semantic errors related to the CATEGORY or degrade the visual quality.\\
\\
Task:\\
Output exactly ONE minimal editing instruction (< 40 words) that achieves the goal derived above. If the PREVIOUS FRAME is provided, use it to maintain subject identity strictly.\\
\\
Output Format (JSON-compatible):
\begin{verbatim}
{
  "instruction": "edit instructions",
}
\end{verbatim}
\end{mybreakablepromptbox}

\begin{mybreakablepromptbox}[Verifier Prompt Template]
You are a visual reasoning verifier for image editing chains.\\

Input Context:\\
- EDITED IMAGE: [new frame]\\
- PROMPT: \{prompt\}\\
- STAGE: \{stage\} (semantic correction OR aesthetic refinement)\\
- DIRECTION: \{direction\} (forward OR backward)\\
- CATEGORY: \{category\} (one of [Attribute Binding, etc.])\\
- PREVIOUS FRAME: [prior frame, optional]\\

Verification Logic:\\
Verify if the transition matches the DIRECTION: if forward, the image must improve in semantics or aesthetics; if backward, it must explicitly degrade or introduce errors. Additionally, check that the main subject identity remains consistent with the PREVIOUS FRAME and no unrelated artifacts appear.\\

Task:\\
- Output SUCCESS(1) if the image clearly executes the intended Forward or Backward goal while preserving unrelated content. \\
- Output FAIL(0) if the change is imperceptible, over-edited, or moves in the wrong direction. \\

Output Format (JSON-compatible):
\begin{verbatim}
{
  "Output": 0 OR 1,
}
\end{verbatim}
\end{mybreakablepromptbox}

\noindent{\textbf{Detail Workflow of UEP.}} 
The UEP follows a simple \emph{"planner-editor-verifier"} iterative workflow.
Given the current frame and the target transition stage, the planner first analyzes the image and produces a concise editing instruction, which is then executed by the editor. The verifier evaluates whether the edited image satisfies the desired stage objective; if not, the process is repeated.

To balance efficiency and precision, the planner and verifier adopt resolution-adaptive perception depending on the transition stage.
For transitions between F1 and F2, the image is resized to 512 × 512, which is sufficient for identifying semantic correctness while reducing computation.
For transitions between F2 and F3, the original 1024 × 1024 resolution is preserved to better assess fine-grained aesthetic details and visual quality. The verifier follows the same resolution strategy.

We set the maximum number of retries to K = 3. If the verifier continues to reject the edited result after exceeding this limit, the pipeline falls back to directly regenerating the image using a strong image generation model (Qwen-Image~\cite{Qwen-Image}) to ensure stable data quality and sample throughput.

\begin{table}[t]
\centering
\caption{\textbf{Performance on Imagine-Bench.}
We report per-type average scores and the overall weighted score.
\textit{Wan2.1 Base} denotes the Wan2.1-T2V-14B base model without CoF fine-tuning.
\textit{Target-only SFT} is fine-tuned using only the final frames of CoF-Evol-Instruct.}
\label{tab:imaginebench}
\setlength{\tabcolsep}{5pt}
\resizebox{\columnwidth}{!}{%
\begin{tabular}{lcccc|c}
\toprule
\textbf{Method}
& \textbf{Attribute Shift}
& \textbf{Hybridization}
& \textbf{Multi-Object}
& \textbf{Spatiotemporal}
& \textbf{Overall} \\
\midrule
Wan2.1 Base
& 5.436
& 6.950
& 5.383
& 6.237
& 5.939 \\
Target-only SFT
& 5.940
& 7.540
& 7.220
& 6.727
& 6.755 \\
\textbf{CoF-T2I (Ours)}
& \textbf{6.969}
& \textbf{8.070}
& \textbf{7.797}
& \textbf{7.287}
& \textbf{7.468} \\
\bottomrule
\end{tabular}}
\end{table}

\section{More Experiment Details}
\label{sec:more_experiment_details}

In this section, we provide additional results on Imagine-Bench and implementation details regarding the continuous video VAE encoding and the system prefix used during training and inference.

\subsection{Additional Results on Imagine-Bench}

\noindent\textbf{Target-only SFT.}
We evaluate the \textit{Target-only SFT} variant on Imagine-Bench to assess the performance when the model is fine-tuned solely on the final output frames, without intermediate reasoning supervision. The per-type average scores are reported in Table~\ref{tab:imaginebench}.

\noindent\textbf{Analysis of the Reasoning Trajectory.}
We further analyze the reasoning trajectory of CoF-T2I on Imagine-Bench by evaluating the generation quality at each step of the chain ($F_1 \rightarrow F_2 \rightarrow F_3$). As shown in Table~\ref{tab:trajectory_imagine}, scores consistently improve across all categories as the reasoning progresses from the initial draft to the final output.

\subsection{Implementation Details}

\noindent\textbf{Continuous Video VAE Encoding.}
Standard causal video VAEs typically adopt a spatiotemporal compression schedule of $1+4n$:
the first frame is encoded without temporal downsampling, while subsequent frames are temporally compressed by a factor of $4$.
As a result, a 3-frame chain $(F_1, F_2, F_3)$ is not directly compatible with the native temporal layout.
To enable continuous video VAE encoding, we pad each training sample to five frames by repeating the final frame:
\[
(F_1, F_2, F_3) \rightarrow (F_1, F_2, F_3, F_3, F_3).
\]
The padded 5-frame clip is then passed through the native video VAE encoder.
During decoding, we only decode and retain the \emph{last frame} as the final output image, while all other decoded frames are discarded.
For text-to-image generation, only the last decoded frame is used as the model output.

\noindent\textbf{System Prompt Prefix.}
To enable the Chain-of-Frame reasoning capability, we append a fixed system prefix to the prompt during both training and inference.
This prefix instructs the model to generate a short refinement chain that preserves concept and composition while improving quality step by step.
The specific prefix used is:

\begin{tcolorbox}[
  colback=gray!5,
  colframe=black!35,
  boxrule=0.5pt,
  arc=2mm,
  left=6pt,
  right=6pt,
  top=5pt,
  bottom=5pt,
  fontupper=\small\ttfamily,
  title=\textbf{System Prompt Prefix}
]
Generate a short refinement chain of the same concept and composition, improving the image step by step.

Prompt: \emph{\textless user prompt\textgreater}
\end{tcolorbox}

\begin{table}[t]
\centering
\caption{\textbf{Analysis of the Reasoning Trajectory on Imagine-Bench.}
We report per-type scores for each frame of the visual reasoning chain, illustrating progressive refinement from the draft to the final output.}
\label{tab:trajectory_imagine}
\setlength{\tabcolsep}{4pt}
\resizebox{\columnwidth}{!}{%
\begin{tabular}{lcccc|c}
\toprule
\textbf{Step}
& \textbf{Attribute Shift}
& \textbf{Hybridization}
& \textbf{Multi-Object}
& \textbf{Spatiotemporal}
& \textbf{Overall} \\
\midrule
Frame 1 (Draft)
& 5.451 & 6.837 & 5.817 & 6.237 & 6.015 \\
Frame 2 (Refine)
& 6.547 & 7.850 & 7.357 & \textbf{7.317} & 7.187 \\
\textbf{Frame 3 (Final)}
& \textbf{6.969} & \textbf{8.070} & \textbf{7.797} & 7.287 & \textbf{7.468} \\
\bottomrule
\end{tabular}}
\end{table}

\begin{figure*}[!t]
    \centering
    \includegraphics[width=0.98\linewidth]{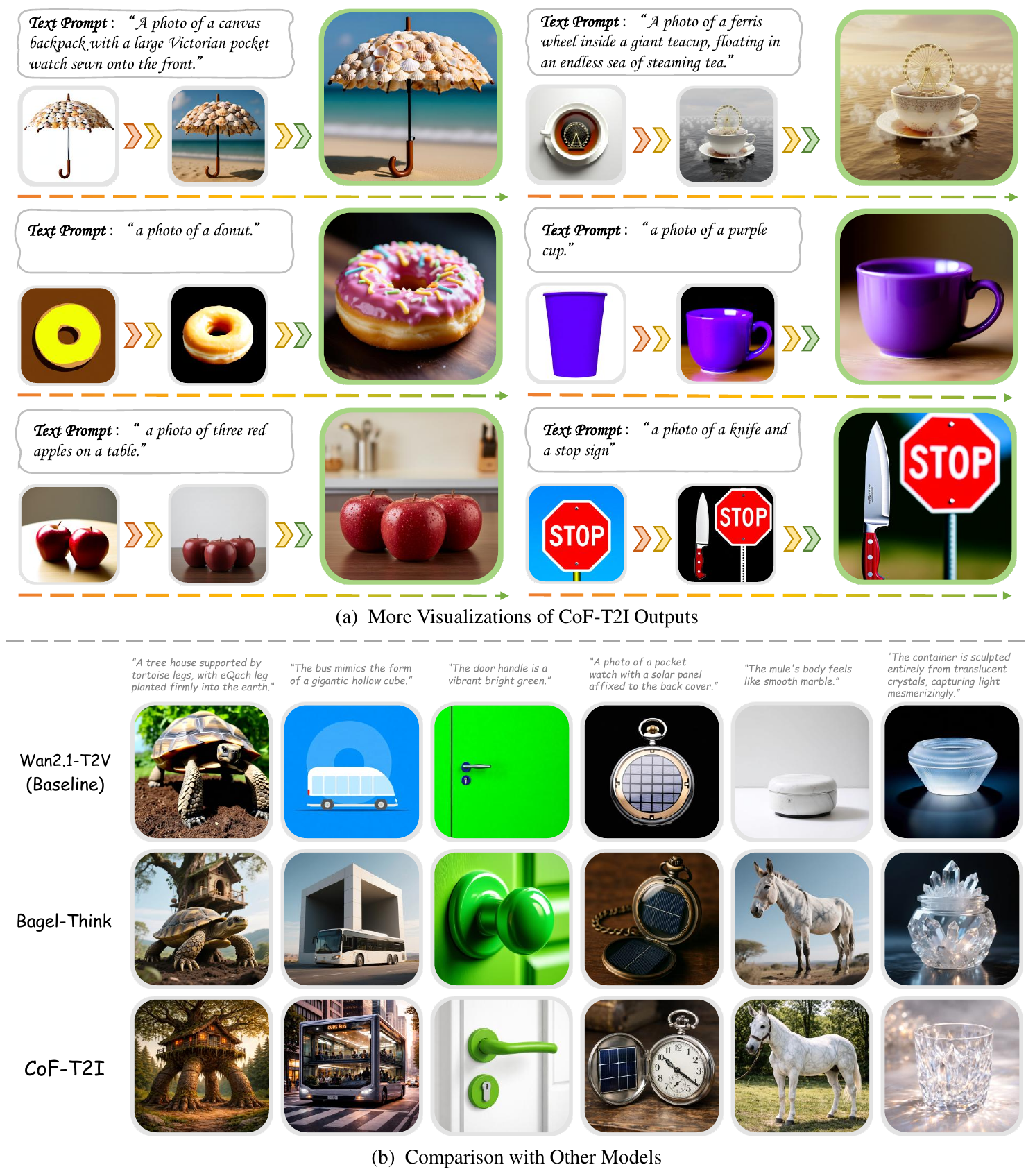}
    \caption{\textbf{Qualitative Analysis: Reasoning Trajectories and Comparisons for \textit{CoF-T2I}}}
    \label{fig:output_fig}
\end{figure*}

\section{Limitations and Future Work} 
\label{app:limit}

While CoF-T2I demonstrates the efficacy of leveraging video models' Chain-of-Frame reasoning for text-to-image generation, our study has not systematically explored its extension to broader task domains, such as video-related tasks (e.g., text-to-video generation) or 3D-related tasks (e.g., text-to-3D synthesis). Extending to text-to-video, for instance, could introduce challenges like handling longer temporal sequences, increased computational demands for multi-frame refinement, maintaining dynamic coherence without introducing unintended motion artifacts, and sourcing high-quality datasets for extended reasoning chains. Additionally, reinforcement learning (RL) techniques, which have proven successful in enhancing Chain-of-Thought reasoning in textual domains through iterative feedback and reward optimization, remain underexplored in our framework; integrating RL with our video-based model could enable more adaptive visual refinements in T2I tasks, potentially improving robustness to diverse prompts and further elevating output quality. Future work will investigate these avenues to unlock the full potential of video foundation models as versatile visual reasoners.

\section{Qualitative Examples}
\label{app:qual}

We provide further qualitative visualizations and comprehensive comparisons in Figure~\ref{fig:output_fig}. In Figure~\ref{fig:output_fig}(a), we showcase the complete reasoning trajectory, including the intermediate frames and the final output alongside their corresponding prompts. The examples encompass diverse generation scenarios, such as imaginative object combination, specific attribute binding, and spatial arrangement. By visualizing the chain-of-frame evolution, we demonstrate how the model iteratively refines semantics or reconstructs details to achieve the final target.

In Figure~\ref{fig:output_fig}(b), we compare our method with the baseline video model, Wan2.1-T2V-14B, and a representative inference-time reasoning model, BAGEL-Think, which interleaves textual Chain-of-Thought with visual generation. As shown, the baseline Wan2.1-T2V tends to rely heavily on training priors and often ignores counter-intuitive instructions; for instance, it generates a standard bus shape instead of the requested "gigantic hollow cube" and fails to render the "vibrant bright green" door handle. While Bagel-Think exhibits better prompt following than the baseline, it still struggles with fine-grained texture synthesis and complex structural deformations, such as the "marble" texture on the mule or the "translucent crystal" container. In contrast, CoF-T2I produces satisfying results with both high photorealistic quality and precise alignment with the prompt, successfully executing challenging instructions that require strong reasoning capabilities.

\end{document}